\documentclass[11pt,a4paper]{article}
\usepackage{times,latexsym}
\usepackage{url}
\usepackage[T1]{fontenc}

\usepackage{graphicx}
\usepackage{multirow}
\usepackage{utfsym}
\usepackage{amsmath}
\usepackage{booktabs}
\usepackage{algorithm}
\usepackage{algorithmic}
\usepackage{tcolorbox}
\usepackage{microtype}

\newcommand{\VarSty}[1]{\textnormal{\ttfamily\color{blue!90!black}#1}\unskip}
\definecolor{ForestGreen}{RGB}{34,139,34}

\usepackage[acceptedWithA]{tacl2021v1}

\title{Generating Visual Stories with Grounded and Coreferent Characters}

\author{Danyang Liu, Mirella Lapata, Frank Keller \\
  Institute for Language, Cognition and Computation \\
  School of Informatics, University of Edinburgh \\
  10 Crichton Street, Edinburgh EH8 9AB \\
  \texttt{danyang.liu@ed.ac.uk, \{mlap, keller\}@inf.ed.ac.uk}}

\date{}

\begin{document}

\maketitle

\begin{abstract}
  Characters are important in narratives. They move the plot forward,
  create emotional connections, and embody the story's themes.  Visual
  storytelling methods focus more on the plot and events relating to
  it, without building the narrative around specific characters. As a
  result, the generated stories feel generic, with character mentions
  being absent, vague, or incorrect. To mitigate these issues,
  we introduce the new task of character-centric story generation and
  present the first model capable of predicting visual stories with
  consistently grounded and coreferent character mentions. Our model
  is finetuned on a new dataset which we build on top of the widely
  used VIST \cite{huang2016visual} benchmark. Specifically, we develop
  an automated pipeline to enrich VIST with visual and textual
  character coreference chains.  We also propose new evaluation
  metrics to measure the richness of characters and coreference in
  stories.  Experimental results show that our model generates stories
  with recurring characters which are consistent and coreferent to
  larger extent compared to  baselines and state-of-the-art
  systems.
\end{abstract}

\section{Introduction}
\label{sec:introduction}

An integral part of storytelling is creating interesting and
believable characters.  In fact, it is not uncommon for writers to
conceptualize their characters visually before crafting a plot around
them.  This character-centric approach enhances narrative coherence,
leading to richer,  engaging, and emotionally resonant stories.  Recent research has explored how character-centric
datasets
\cite{porvatov-etal-2024-big,chen2023large,brahman-etal-2021-characters-tell}
as well as different representations of characters might contribute to
the automatic analysis and generation of narratives
\cite{gurung2024chiron,inoue2022learning,bamman-etal-2013-learning,
  kim_learning_2018,liu2020character,brahman-etal-2021-characters-tell,
  chen_persona_2024, li_zero-shot_2024, xu_character_2024,
  yu_few-shot_2022, yang_doc_2023, yang_re3_2022}.

\begin{figure*}[t]
    \centering
    \includegraphics[width=\linewidth]{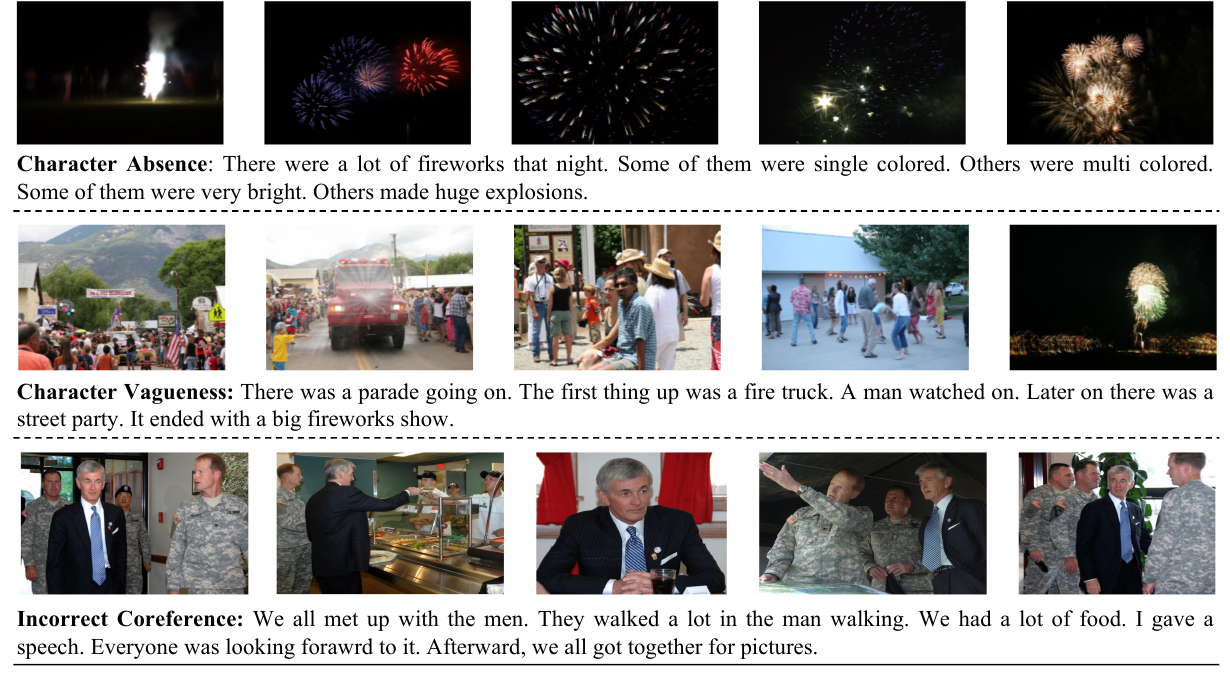}
    \caption{{Examples from the VIST dataset illustrating how the absence of characters or vague character references affect story coherence and engagement. The first story is purely descriptive without any characters, lacking emotional depth and narrative engagement. No human presence means no perspective, making it static and impersonal. In the second story, while a character is mentioned (``a man''), he adds nothing to the story. The man is passive, disconnected from events, and does not drive the narrative, making the story feel just as flat as the first one. The third story fails to refer to the protagonist correctly, switching between ``we'', ``they'', and ``I'', which causes the story to be confusing and illogical.}}
    \label{fig:examples-character-influence}
\end{figure*}

The importance of characters seems to go unnoticed in narrative tasks
spanning multiple modalities such as visual storytelling, which
involves narrating a story based on a sequence of images. Existing
approaches focus on detecting objects in images and discovering
relationships between them; characters are treated the same as other
objects, without any special consideration in the generation process.
For example, the most popular approaches to visual storytelling
\cite{wang-etal-2024-sco,chen2021commonsense,hsu2020knowledge,
  yang2019knowledgeable} exploit external knowledge bases to enrich
and link detected concepts.  While such methods can describe a
coherent sequence of events, they fail to effectively ground character
mentions to their visual depictions and generate character-centric
stories.  As a result, character mentions are often absent, vague
(e.g.,~rendered with plural references such as ``they'' or ``we''), or
incorrect, and the stories feel generic, lacking in detail and
context.
{Figure~\ref{fig:examples-character-influence} gives representative examples
that illustrate how story quality depends on the presence of visually grounded characters.}

\begin{figure*}[t]
    \centering
    \includegraphics[width=\textwidth]{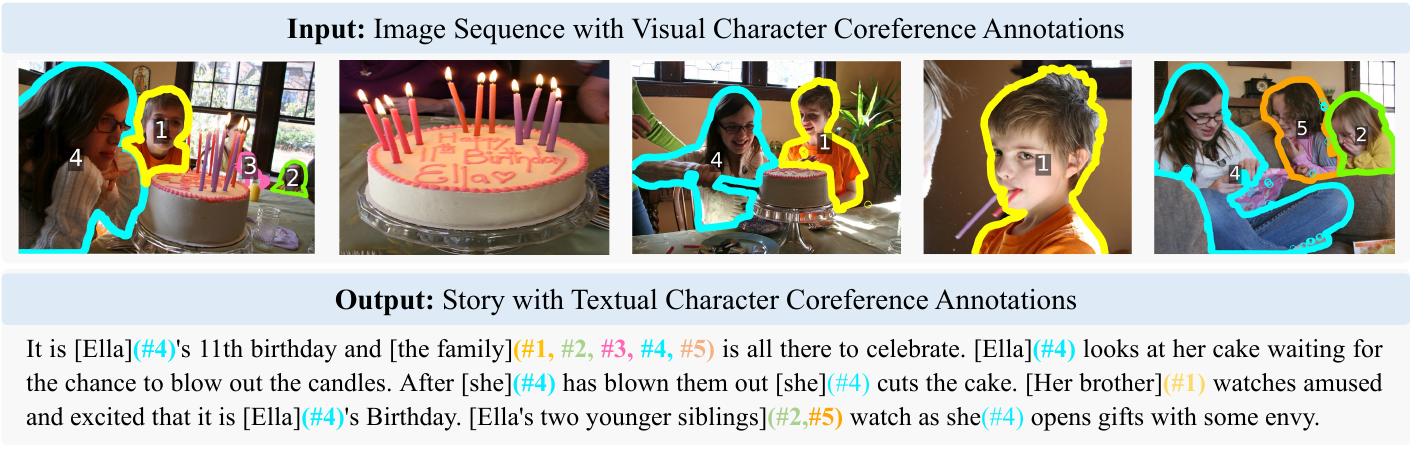}
    \caption{A sample from the VIST dataset \cite{huang2016visual}
      augmented with character chains. Visual characters are outlined
      by segmentation mask boundaries, where same color indicates same
      characters. Each character bears a unique label overlaid in the
      center of the mask. Output stories are annotated with textual
      character chains, and each character mention is aligned to
      visual characters (e.g.,~\mbox{(\#4)} refers to visual
      segment~4).}
    \label{fig:model_input_output}
\end{figure*}

It is not possible to create character-centric stories without knowing
who the characters are. We therefore propose to identify them in an
image sequence together with their visual coreference chains.  In
Figure~\ref{fig:model_input_output}, different characters are
segmented using different colors, and are assigned a unique
ID. Characters with the same ID in different images form a coreference
chain. There are five characters in the image sequence (labels~1--5),
and four coreference chains (characters~1 and~4 are shown in three
images, character 2~is shown in two, while characters~3 and 5~are
depicted only once). We would then expect a generation model
that takes these annotations into account to refer to the same
character consistently, creating textual coreference chains which are
nevertheless grounded. Again, in Figure~\ref{fig:model_input_output}
``Ella'' and ``she'' refer to visual character~4, whereas ``Ella's two
younger siblings'' refers to characters~2 and~5. Note that without
forcing the model to explicitly ground character mentions in 
images, there is no way of knowing which visual characters the model
is talking about.


In this work, we introduce the task of \emph{character-centric} visual
story generation, which aims to create stories with character mentions
grounded to the input image sequence. As explained earlier, the task
requires understanding coreference relationships between visual
characters and aligning textual character mentions to their visual
counterparts. This type of grounding is key to generating coherent and
detailed stories, anchoring textual mentions to specific visual
appearances, and keeping the story closely linked to the image
context.

Perhaps unsurprisingly, existing datasets do not contain annotations
such as those shown in Figure~\ref{fig:model_input_output}. We augment
the widely used VIST dataset \cite{huang2016visual} with visual and
textual character annotations following a fully automated pipeline. We
use \textsc{Otter-LLaMA-7B} \cite{li2023mimic}, which has been
instruction-tuned for multiple image understanding, as our backbone
model and train it on image sequences and stories enriched with visual
and textual coreference chains (see
Figure~\ref{fig:model_input_output}).  We render visual character
chains into visual prompts and enforce character grounding in the
generated stories via a new format which links character mentions to
visual ids. We further introduce novel automatic metrics which assess
character richness and coreference in stories, and propose to use LLMs
as evaluators (LLM as a judge) to perform side-by-side comparisons of
stories generated by different systems. The contributions of our work
can be summarized as follows:

\begin{itemize}
\item We introduce the new task of character-centric story generation,
  and present the first model capable of generating stories with
  consistent and grounded character mentions.

\item Recognizing the lack of character annotations in visual stories,
  we enrich the VIST benchmark \cite{huang2016visual} with visual and
  textual character coreference chains and their corresponding
  alignment.  The new benchmark, which we call VIST++, contains
  detailed annotations for 300K unique characters over 40K visual
  stories.
    
\item We propose new evaluation methods to measure the richness of
  characters and coreference in stories. We further replace costly
  human evaluation with an LLM-as-a-Judge approach, which we use to
  compare visual stories along various dimensions (e.g.,~specificity,
  coherence, grounding).
\end{itemize}


\section{Character-centric Visual Story Generation}
\label{sec:approach}

Our work aims to address limitations of current visual storytelling
systems, which struggle to accurately recognize character coreference
relationships in the input images.  Different from traditional visual
storytelling tasks, where the input is an image sequence and output is
a story, our character-centric approach takes image sequences and their
corresponding visual character coreference chains as input, and
produces a story with grounded and coreferring characters as output.
By \emph{explicitly} linking textual characters to their visual
counterparts, we ensure character consistency, reduce vague or
incorrect references, and facilitate the generation of more accurate,
detailed, and engaging narratives. In the following, we first discuss
how we automatically augment VIST \cite{huang2016visual} with
character chains (Section~\ref{sec:dataset_creation}) and then present
our visual story generation model (Section~\ref{sec:model}).

\subsection{The VIST++ Dataset}
\label{sec:dataset_creation}


We developed an automated pipeline to enrich VIST
\cite{huang2016visual} with detailed character annotations, including
fine-grained character labels in images and aligned textual
coreference chains. The resulting VIST++ dataset contains~40K visual
stories, with 300K~unique characters, each associated with
segmentation masks across images and textual coreference chains,
amounting to a total of 520K character appearances.


Our annotation pipeline includes three subtasks: (1)~visual
character coreference, i.e.,~identifying characters in the image
sequence and grouping those that are the same person into 
coreference chains; (2)~textual character coreference involves
detecting character mentions and identifying coreference chains in
the story text; and (3)~multimodal alignment links textual to visual
coreference chains, yielding multi-modal chains.  We next introduce
our automatic procedure for each task.

\begin{figure*}[t]
    \centering
    \includegraphics[width=\textwidth]{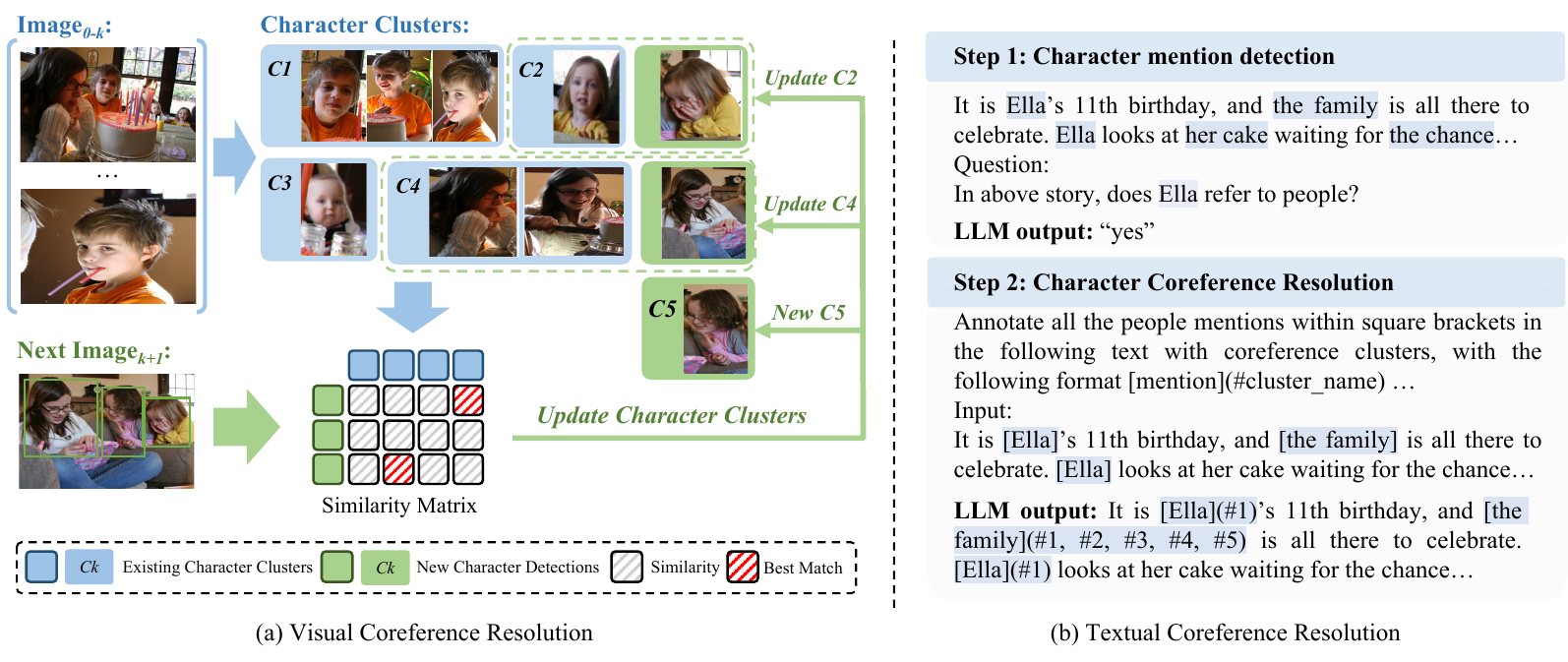}
    \caption{Illustration of the incremental clustering algorithm for
      creating visual coreference chains and prompts used in textual
      coreference resolution. (a)~Character detections in
      Image$_{k+1}$ are compared against character clusters from the
      previous $k$~images to generate a visual similarity matrix. Best
      matching character detections are added to existing
      clusters. Detections that do not match any clusters are grouped
      into new clusters. (b)~A QA-based prompting method is used to
      identify entities referring to characters (Step~1). Then
      character clusters are identified using a structured prompt
      template, which can handle singular and plural character
      mentions (Step~2).}
    \label{fig:coref_in_one}
\end{figure*}

\subsubsection{Visual Character Coreference}
\label{sec:visual_coref}


To obtain visual character chains, we first detect characters in an
image sequence and then identify which detections represent the same
character, thereby constructing coreference chains.

Previous visual coreference methods
\cite{schroff2015facenet,liu2023detecting} group bounding boxes
(e.g.,~based on facial features) directly into a fixed number of
clusters which should ideally vary from image to image. Our
approach differs in two respects. Firstly, we use segmentation masks
rather than bounding boxes; the latter are well suited to rectangular
shapes, but are less accurate with objects that have irregular
boundaries. Moreover, segmentation masks can distinguish between
multiple overlapping objects -- in our case images can contain
densely populated background characters. 
Secondly, we propose to cluster segmentation masks in chains following
an incremental algorithm (shown in Figure~\ref{fig:coref_in_one}a)
which does not require the number of clusters to be known in advance
and does not depend on facial features, which often fail to detect
side-views of faces (in visual stories).


Given an image sequence, we first apply a pretrained object detector
and retain detections with the label \textit{Person} and a confidence
score higher than~0.9 which cover at least 10\% of the image (to
remove background characters). Visual chains are constructed
incrementally: First, all characters detected in the first image are
considered new and added to the chain.  Subsequently, for every new
character detected in image$_{k+1}$, we compute the pairwise
similarity between this character and clusters from previous
images$_{0:k}$, thus obtaining a similarity matrix. We construct a
bipartite graph and the maximum weighted bipartite matching is
computed using the Hungarian algorithm \cite{kuhn1955hungarian}. Nodes
in this bipartite graph represent crops (character instances cropped
using their segmentation masks), while edge weights correspond to
visual similarities between crops.  The algorithm finds a one-to-one
alignment that maximizes the sum of the similarities between crops.

We added a threshold condition to the Hungarian algorithm which allows
to create new clusters if new characters are not visually similar to
existing ones.  Specifically, if a character detection does not result
in a match with a similarity value higher than a threshold, it is
added as a new character to the visual coreference chains.  An
illustration of our algorithm is shown in
Figure~\ref{fig:coref_in_one} and a more formal description in
Appendix~\ref{app:sequential_matching}.  We employed
DETR~\cite{carion2020end} to detect character bounding boxes, and SAM
\cite{kirillov2023segment} to obtain segmentation masks for them.  We
used OpenCLIP ViT-G/14~\cite{ilharco_gabriel_2021_5143773} for visual
feature extraction and measured the visual similarity between two
crops with \textsc{LLaVA1.5-13b}~\cite{liu2024llavanext} (i.e.,~by
providing two crops and asking the model whether they represent the
same character).


\subsubsection{Textual Character Coreference}
\label{sec:textual_coref}

Current state-of-the-art coreference resolution models generally
struggle to effectively handle the coreference of plural
nouns~\cite{liu2023detecting,le2023large,hicke2024lions}, which are
very common in stories.  Our proposed coreference resolution method is
based on LLMs (see Figure~\ref{fig:coref_in_one}b); through in-context
learning, we are able to detect coreference relationships for
characters denoting a single \emph{and} multiple entities
(i.e.,~plural and collective nouns). More specifically, our method
consists of two steps:

\paragraph{Character mention detection} Given a visual story, we
detect all entities mentioned therein using spaCy \cite{spacy2023} and
mark these with square brackets, denoting a span to annotate in the
next step (e.g.,~James $\rightarrow$ [James]). We identify
character-specific entities with a zero-shot prompting method which
takes the story as input and spaCy entities and identifies via
question-answering whether these denote persons (either groups or
singular entities). Our full prompt is provided in the Appendix (see
Table~\ref{tab:prompt_character_detection}) and an illustration in
Figure~\ref{fig:coref_in_one}b (Step~1).


\paragraph{Character coreference resolution}

Using in-context learning, we annotate character spans with cluster
IDs (e.g.,~[James] $\rightarrow$ [James](\#1)). We use
five examples, and a prompt that instructs the LLM to annotate
mentions within square brackets with markdown tags to indicate
clusters, resulting in the format [mention](\#cluster\_name). Note
that plural mentions can refer to more than one characters (e.g., [We]
$\rightarrow$ [We](\#1, \#2)).  For an illustration, 
see Figure~\ref{fig:coref_in_one}b (Step~2); the full prompt is 
given in the Appendix (see
Table~\ref{tab:prompt_coreference_resolution}).  We used LLAMA3-70B
\cite{touvron2023llamaopenefficientfoundation} to identify character
mentions and obtain textual chains.

\subsubsection{Multimodal Character Alignment}
\label{sec:alignment}

Finally, we align textual and visual coreference chains.  We again
model this alignment as a bipartite graph matching problem.
Specifically, we create a matrix where each cell represents the
similarity between a textual and visual chain and apply the Hungarian
algorithm to find the best match.  We measure chain similarity
following the method proposed in \citet{liu2023detecting} which
compares the distribution of characters in images with their
distribution in sentences. For instance, if a character is mentioned
in the first and second sentence, and there are two coreferent visual
detections in the first and second image, then these are likely to
refer to the same character.


More formally, a $C$-dimensional binary vector (where~$C$ is
the number of images or sentences\footnote{We assume the number of
  sentences in a story is the same as the number of input images.})
 represents the distribution of a character. Let $C_i$~denote the
$i$-th textual/visual character, then $C_i[k]=1$ refers to the $i$-th
character being present in the $k$-th~sentence/image. The similarity
between a visual  and atextual character is computed as the
dot product of their respective binary vectors.

\subsubsection{Quality Evaluation}
\label{sec:quality-evaluation}


We evaluated our automatic pipeline against VIST-Character
\cite{liu2023detecting}, a high-quality, human-annotated dataset
containing rich character annotations for 770 visual stories from the
VIST \cite{huang2016visual} test set, including visual and textual
coreference chains and their alignments. We further split this dataset
into 300~stories for validation and 470~stories for testing. Results
on the test set are shown in Table~\ref{tab:detec-coref-results}. We
report the precision and recall of automatically detected
characters. We assume a textual detection is correct if the head of
the noun phrase in question is the same as the gold-standard one. For
visual detections, a predicted region is considered correct when the
Intersection over Union (IoU) \cite{yu2016unitbox} is higher than a
threshold of~50\%. We evaluate coreference chains using the B$^3$
metrics~\citet{cai2010evaluation}, which indicate the average
percentage of correctly detected mentions in a chain. Specifically, we
compute precision and recall for each individual mention (see Appendix
for a formal definition).

\begin{table}[t]
\centering
\begin{tabular}{lcccc}
\hline
\multicolumn{1}{c}{} & \multicolumn{2}{c}{Detection} & \multicolumn{2}{c}{Coreference} \\
\multicolumn{1}{l}{\multirow{-2}{*}{Model}} & P & R & P & R \\
\hline
Visual & {41.8} & {78.4} & {70.4} & {74.2} \\ 
Textual &{84.9} & {94.5} & {77.2} & {80.3} \\
Multimodal & --- & --- & 39.7 & 42.5 \\ 
\hline
\end{tabular}
\caption{Results (in~\%) for character detection and coreference chain
  identification. P/R are precision/recall.
}
\label{tab:detec-coref-results}
\end{table}

In general, we observe good precision and recall for textual character
detection and coreference. Visual characters are harder to detect
(precision 41.8\%), but we are able to build accurate coreference
chains for the characters we identify. Multimodal alignment is harder
precisely because of visual detection being challenging. Nevertheless
we obtain a 10\%-15\% improvement in multimodal chain accuracy over
\citet{liu2023detecting} who use simple, unsupervised models for this
task based on distributional similarity and CLIP embeddings
\cite{radford2021learning}. We provide a more detailed comparison in
Appendix~\ref{app:pipeline_anylasis}.

\begin{table}[t]
\centering
\resizebox{.48\textwidth}{!}{%
\begin{tabular}{lccrrr}
\hline
Dataset & Auto. & Coref. & \#Stories & \#Chars & \#Boxes \\
\hline
VIST-Char  & \textcolor{red}{\usym{2717}} & \textcolor{ForestGreen}{\usym{2713}} & 0.7K & 3K & 5K \\
VWP & \textcolor{red}{\usym{2717}} & \textcolor{red}{\usym{2717}} & 12K & 157K & 157K \\
VIST++ & \textcolor{ForestGreen}{\usym{2713}} & \textcolor{ForestGreen}{\usym{2713}} & \textbf{40K} & \textbf{300K} & \textbf{520K}\\
\hline
\end{tabular}%
}
\caption{Statistics for datasets with character annotations. `Auto'
  indicates whether the dataset has been annotated
  automatically. '\#stories' and '\#chars' refer to the number of
  visual stories and unique characters.}
\label{tab:stat}
\end{table}

Finally, in Table~\ref{tab:stat} we compare VIST++ to existing
character-centric visual story generation datasets.  VIST-Character
\cite{liu2023detecting} has been manually curated but is a relatively
small-scale, mainly designed for evaluation.  Visual Writing Prompts
(VWP; \citealt{hong-etal-2023-visual-writing}) is larger with
12K~stories and character annotations provided by humans. However, it
does not include coreference chains and consists of images selected
from movie scenes. VIST++ contains the highest number of stories and
silver-standard annotations; although we focus on VIST, our automatic
pipeline could be applied to related datasets such as VWP.

\subsection{Visual Story Generation Model}
\label{sec:model}

Our generation model is trained on VIST++. It is able to identify
characters across multiple images and generate stories with multimodal
character chains. We leverage the visual understanding and text
generation capabilities of Large Vision-Language models
(LVLMs). Specifically, we employ \textsc{Otter} \cite{li2023mimic} as
our backbone model, which has been instruction-tuned for multiple
image understanding.  \textsc{Otter} \cite{li2023mimic} is the
instruction-tuned version of
OpenFlamingo~\cite{awadalla2023openflamingo} (an open-source
replication of DeepMind's Flamingo models). Openflamingo comprises a
LLaMA-7B \cite{touvron2023llamaopenefficientfoundation} language
encoder and a CLIP ViT-L/14 \cite{radford2021learning} vision
encoder. The two modalities are interleaved through gated
cross-attention layers that allow the model to combine information
from the visual and textual streams.

\paragraph{Training}

We finetune \textsc{Otter} \cite{li2023mimic} on pairs of image
sequences and their corresponding stories. In VIST++, characters are
outlined with segmentation masks, where same color indicates same
characters, i.e,~a coreference chain (see
Figure~\ref{fig:model_input_output}). Characters are further labeled
with a unique ID number, overlaid in the center of the mask, and
characters with the same ID in different images are assumed to be the
same person. These visual annotations or marks \cite{yang2023set}
serve as visual prompts to \textsc{Otter}.


During training, the model learns to predict a story with character
grounding and coreference information.  Aside from visual prompting,
grounding is facilitated by training the model to \textit{verbalize}
textual character chains and their grounding. The model learns to
predict textual chains, i.e.,~it marks character mentions with a
special symbol, and is explicit about which visual segmentation they
refer to (e.g.,~[Ella](\#4) refers to visual segment~4, [she](\#4)
refers to Ella and segment~4, whereas [We](\#1, \#2) refers to
segments~1 and 2).

Within \textsc{Otter}, we freeze the parameters of the language
encoder (\textsc{LLaMA-7B}) and the vision encoder (\textsc{CLIP
  ViT-L/14}), and only update the parameters of the Perceiver
\cite{jaegle2021perceiver} resampler module, the cross-attention
layers added to the language encoder, and the input/output embeddings
of the language encoder. This results in approximately 1.44~billion
trainable parameters.

\paragraph{Inference}

During inference, the input to the model is a regular image sequence
(without any character related annotations). We use the automated
pipeline described in Section~\ref{sec:visual_coref} to obtain visual
character chains which are in turn converted into visual prompts,
following the same method used during training. The image sequence
with the visual character prompt serves as input to our model which
is trained to generate a story with character grounding and
coreference information.



\section{Experiments}
\label{sec:experiments}


\subsection{Datasets}

We perform experiments on VIST \cite{huang2016visual}, which is the
most widely used visual storytelling dataset; it contains
10,117~Flickr albums and 210,819~unique photos.  Each training sample
consists of~$k=5$ images and a corresponding story of~$k=5$
sentences. Training, validation, and test sets contain 40,155, 4,990,
and 5,055 unique stories, respectively. Our model is trained on
VIST++, making use of the character-centric annotations discussed in
Section~\ref{sec:dataset_creation}. We report results on the original
VIST test set (no annotations) and also on the VIST-Character subset
\cite{liu2023detecting} with gold-standard annotations when evaluating
character specific properties.

{The Visual Writing Prompts (VWP,~\citealt{hong-etal-2023-visual-writing}) dataset is a recently introduced benchmark for character-grounded story generation with curated image sequences.
However, it is not entirely suited to our task, as it only annotates the bounding box of a character upon their first appearance without providing a coreference chain, preventing us from evaluating the character coreference.
Nevertheless, we evaluate our models, trained on the VIST dataset, on VWP to demonstrate their generalization capability. We use metrics other than character coreference.
}

\subsection{Implementation Details}

We finetuned \textsc{Otter} with a learning rate of 1e-5, batch size
of 32, and warm-up ratio of 0.05.  During inference, we employed
greedy decoding.  We accelerate the training process using the
DeepSpeed framework~\cite{rasley2020deepspeed}.

\subsection{Evaluation}
\label{sec:evaluation-metrics}

Many previous studies
\cite{liu2023visual,hsu2021plot,hsu2020knowledge,hu2020makes,
  yang2019knowledgeable,modi2019steep} have highlighted the
limitations of metrics based on lexical matching for story evaluation.
They correlate poorly with human judgments and do not effectively
measure semantic similarity with human-written stories or the lexical
richness of the generated stories.  In this work, we employ
\textit{story-specific} metrics to assess various aspects of story
quality, including diversity, grounding, naturalness, and
readability. We also propose automatic metrics that assess character
richness and the accuracy of coreference chains. Finally,
we introduce LLM-as-Judge evaluators
\cite{zheng2024judging,liu2024aligning,liusie2023zero} to perform
binary comparisons of stories generated by different systems.

\paragraph{Diversity} We use Inter-story Repetition
\cite{yao2019plan,goldfarb-tarrant-etal-2020-content}, which examines
trigram repetition across stories to measure diversity.  High
inter-story repetition suggests that the model tends to generate the
same story even when conditioned on different image sequences.

\paragraph{Grounding} We use GROOViST \cite{surikuchi2023groovist} to
assess whether the stories accurately represent the content of the
image sequences. GROOViST employs CLIPScore \cite{hessel2021clipscore}
to compute the similarity between noun phrases in the story and
bounding boxes in the images, producing an average score which favors
stories with concrete words as they are more likely to be visible.

\paragraph{Naturalness} We use MAUVE
\cite{pillutla-etal:mauve:neurips2021} to measure the naturalness of
the stories. MAUVE has a high correlation with human judgments and
computes the similarity between the distributions of human- and
machine-generated texts.

\paragraph{Reading difficulty} We employ the Flesch-Kincaid Grade
Level (FKGL)~\cite{kincaid1975derivation} to measure the reading
difficulty of stories. FKGL is a well-established readability formula
used for text quality assessment:
\begin{equation}
    \textrm{FKGL} = a \frac{N_{word}}{N_{sentence}} + b \frac{N_{syllable}}{N_{word}}+c
\end{equation}
where $a$~is 0.39, $b$~is 11.8, and $c$~is $-$15.59 as defined in~\citet{kincaid1975derivation}.
The FKGL values denote reading age, ranging from~0 to~18, So, lower values suggest the text is easier to read and higher values
denote increased reading difficulty.

\paragraph{Character richness and coreference} We report the number of
characters and character mentions in generated stories as measures of
character richness.  We also evaluate the accuracy of multimodal
coreference chains using the B$^3$ score
\cite{cai-strube-2010-evaluation}. We only use B$^3$ precision, as we
are more interested in the correctness of mentions within a predicted
chain rather than ensuring that every gold-standard visual character
is mentioned. For comparison systems which do not detect characters or
ground them to images, we employ the pipeline described in
Section~\ref{sec:textual_coref} on their generated output. Note that
character metrics are computed against the VIST-Character
\cite{liu2023detecting} test set, as it contains manually annotated
visual coreference chains.



\paragraph{LLM as a judge} We further introduce an LLM-based evaluator
to perform side-by-side comparisons of system output. Specifically, we
employ human side-by-side judgments of visual stories from previous
work \cite{hsu-etal-2022-learning,liu2023visual} covering the
following criteria: Specificity, Coherence, Engagement, Grounding,
Characters, and Overall Preference. We then designed a prompt
targeting each dimension of story quality (see
Table~\ref{tab:prompt_quality_estimation2} in the Appendix) then and
evaluated LLM accuracy against human judgments.

\begin{table}[t]
\centering
\resizebox{.48\textwidth}{!}{%
\begin{tabular}{lcccccc}
  \toprule

  \multicolumn{1}{c}{Input} & SPE & COH & ENG & GRD & CHA & OVR \\ \midrule
  Multimodal & 81.24 & 75.26 & 74.23 & 74.52 & 70.52 & 75.26\\
  Text-Only & 78.10 & 71.36 & 75.19 & --- & 65.11 & 74.43\\ \bottomrule
\end{tabular}%
}
\caption{Accuracy of \textsc{GPT-4o} against human pair-wise judgements of story quality. SPE, COH, ENG, GRD, CHA, OVR
  are short for specificity, coherence, engagement, grounding,
  characters, and overall preference.} 
\label{tab:llm-judge-agreement}
\end{table}

Table~\ref{tab:llm-judge-agreement} reports the accuracy of
\href{https://openai.com/index/hello-gpt-4o/}{\textsc{GPT-4o}} as a
judge (more details in Appendix~\ref{app-llmjudge}). We present results for
LLMs like \textsc{GPT-4o}, which can process images
\emph{and} text, but also simulate the use of text-only LLMs by giving
\textsc{GPT-4o} the story without the associated image
sequence. \textsc{GPT-4o} agrees with
human judgments 70--80\% of the time across the story quality
dimensions.  This suggests that it is a fairly accurate evaluator and
could replace costly and time-consuming human judgments.

{\paragraph{dHM}
A concurrent study by \citet{surikuchi2024not} introduces dHM, a human-centric evaluation framework for model-generated stories, focusing on key dimensions relevant to visual story generation. This framework incorporates three reference-free evaluation metrics: GROOViST~\cite{surikuchi2023groovist} for visual grounding, RoViST-C for coherence, and RoViST-NR for non-redundancy/repetition.
We incorporate these metrics into our analysis to ensure a more comprehensive and up-to-date evaluation.}


%


\section{Results}

\begin{table*}[t]
\centering
\resizebox{\textwidth}{!}{
\begin{tabular}{l|ccc|ccccc|cccc}
\toprule
& \multicolumn{3}{c|}{Character Metrics ($\uparrow$)} &
\multicolumn{5}{c|}{Story Metrics} &
\multicolumn{4}{c}{N-gram-based Metrics ($\uparrow$)} \\ 
\multicolumn{1}{l|}{Model} & \multicolumn{1}{c}{\#CHAR} & \multicolumn{1}{c}{\#MENTS} &
COREF &  \#Words & {DIV}
($\downarrow$) & {GRD} ($\uparrow$) &
{MAU} ($\uparrow$) & {RDD} & B-4 & RL & MET & CID\\ 
\toprule
MCSM &  1.21 & 2.01 & 10.43 & 39.11 &  77.48  & 48.12 & 11.01  & 5.30 & 8.1 & 27.7 & 31.4 & 7.6 \\
Iter-Blueprint  & 1.52 & 2.13 & 12.56 & 38.67 & 72.70 & 62.54  & \textbf{28.25} & 6.10 & 7.0 & 26.1 & 30.3 & 5.5\\ 
\textsc{SOtter}  & 1.91 & 3.15 & 26.61 & 50.18 & 56.10 & \textbf{65.97} & 10.68 & 5.12 & 5.4 & 18.2 & 23.1 & 2.9 \\ 
\textsc{SOtter}++ & \textbf{2.98} & \textbf{4.98} & \textbf{32.16} & 51.75
& \textbf{54.58} & 64.70 & 10.47 & 5.42 &5.1 & 18.0 & 22.6 & 2.6 \\ 
\midrule
GPT-4V (0-shot) & 5.99 & 9.76 & 29.96 & \hspace*{-.2cm}118.34
& \hspace*{.2cm}9.61 & 58.30 & \hspace*{.2cm}2.41 & \hspace*{-.2cm}12.86 & 0.8 & 15.2
& 15.7 & 0.1\\
{GPT-4V (<50 wds.)}  & {1.98} & {3.99} & {30.51} & {56.24}
& {\hspace*{.2cm}8.00} & {61.68} & {\hspace*{.2cm}2.40} & {9.89} & {1.7} & {14.4}  
& {16.6} & {5.0} \\
{GPT-4V (5-shot)}  & {2.55} & {4.23} & {29.98} & {74.57}
& {\hspace*{-.1cm}15.30} & {48.15} & {\hspace*{.2cm}5.18} & {\hspace*{-.2cm}10.95} & {1.5} & {13.5} 
& {16.5} & {0.6}\\
\midrule
Gold & 4.11 & 5.09 & 51.36 & 48.95 & 37.24 & 73.60 & --- & 5.51 & --- & --- & --- & --- \\
\bottomrule
\end{tabular}%
}
\caption{Automatic evaluation results. We report character-centric
  metrics, story-specific metrics, and metrics measuring lexical
  similarity between system stories and their references. The
  abbreviations \#CHAR, \#MENTS, COREF correspond to the number of
  characters, mentions, and the coreference score. DIV, GRD, MAU, RDD,
  B-4, RL, MET, and CID are short for Diversity (lower is better),
  Grounding, MAUVE, Reading Difficulty, Blue-4, RLSum, METEOR, and
  CIDER, respectively. Best results, excluding GPT-4V, are highlighted
  in bold.}
\label{tab:generation_results}
\end{table*}

\begin{table}[h]
\centering
\resizebox{.5\textwidth}{!}{%
\begin{tabular}{lccc}
\hline
{Model} & {GROOViST} & {RoViST-C} & {RoViST-NR} \\ \hline
{MCSM} & {48.12} & {0.67} & {0.90} \\
{Inter-Blueprint} & {62.54} & {0.72} & {0.93} \\
{\textsc{SOTTER}} & {65.97} & {0.78} & {0.95} \\
{\textsc{SOTTER}++} & {64.70} & {0.80} & {0.95} \\
\hline
\end{tabular}%
}
\caption{{Evaluation results based on the human-centric metrics proposed by \citet{surikuchi2024not}. Higher scores indicate better performance in visual grounding (GROOViST), coherence (RoViST-C), and non-redundancy (RoViST-NR).}}
\label{tab:aaadditional_results}
\end{table}

\begin{table*}[t]
\centering
\resizebox{\textwidth}{!}{
\begin{tabular}{l|ccc|ccccc|cccc}
\toprule
& \multicolumn{3}{c|}{{Character Metrics ($\uparrow$)}} &
\multicolumn{5}{c|}{{Story Metrics}} &
\multicolumn{4}{c}{{N-gram-based Metrics ($\uparrow$)}} \\ 
\multicolumn{1}{l|}{{Model}} & \multicolumn{1}{c}{{\#CHAR}} & \multicolumn{1}{c}{{\#MENTS}} &
{COREF} &  {\#Words} & {DIV
($\downarrow$)} & {GRD ($\uparrow$)} &
{MAU ($\uparrow$)} & {RDD} & {B-4} & {RL} & {MET} & {CID}\\ 
\toprule
{MCSM} &  - & - & - & - &  -  & - & - & - & - & - & - & - \\
{Iter-Blueprint}  & {1.66} & {2.55} & - & {57.34} & {71.11} & {54.34}  & {\textbf{22.25}} & {8.52} & {6.1} & {25.1} & {31.1} & {5.2}\\ 
{\textsc{SOtter}}  & {2.01} & {3.22} & - & {51.54} & {57.12} & {59.47} & {9.06} & {8.01} & {5.6} & {18.6} & {23.0} & {2.7} \\ 
{\textsc{SOtter}++} & {\textbf{2.97}} & {\textbf{4.88}} & - & {52.04}
& {\textbf{53.48}} & {\textbf{61.21}} & {10.87} & {8.16} & {5.7} & {18.9} & {22.6} & {2.6} \\ 
\bottomrule
\end{tabular}%
}
\caption{{Evaluation results on the VWP dataset for models trained on VIST to show the generalization ability of our models.}}
\label{tab:generation_results-VWP}
\end{table*}

Our experimental results are summarized in
Table~\ref{tab:generation_results}.  The first rows present the
performance of state-of-the-art storytelling systems from the
literature.  \textbf{MCSM}~\cite{chen2021commonsense} exploits a
commonsense knowledge graph to represent concepts depicted in the
images and their relations, and uses a \textbf{M}aximal
\textbf{C}lique \textbf{S}election \textbf{M}odule to identify which
ones to write a story about. MCSM uses BART-large \cite{lewis2020bart}
to generate the story based on selected concepts (and image features).
\textbf{Iter-Blueprint}~\cite{liu2023visual} is built on top of
BART-base and leverages a sequence of question-answer pairs as a
blueprint (story plan) for selecting salient visual concepts and
determining how to present them in the story. The model employs an
incremental generation strategy, gradually constructing the blueprint
and its corresponding story sentence-by-sentence.

\textbf{\textsc{SOtter}++} is our story generation model based on
\textsc{Otter} and finetuned on the proposed VIST++ dataset, whereas
\textbf{\textsc{SOtter}} is finetuned on vanilla VIST and does not
have any specific knowledge about characters or their grounding.
Finally, Table~\ref{tab:generation_results} includes results for
\textbf{GPT-4V}\footnote{https://openai.com/index/gpt-4v-system-card/} (zero-shot setting, prompt in Figure~\ref{tab:prompt_gpt4v}) which is one of the most
performant multimodal LLMs.
All models in Table~\ref{tab:generation_results} were evaluated on the
same VIST test set.


\paragraph{\textsc{SOtter}++ leads on character-centric metrics.}  The
number of characters and their mentions in stories generated by
\textsc{SOtter} is similar to that of previous visual storytelling
systems.  However, the number of characters and frequency of character
mentions increase substantially for \textsc{SOtter}++, which is
trained on VIST++. In addition, the accuracy of the coreference chains
improves by more than~5\%.  This indicates that VIST++ effectively
enables LVLMs to generate stories with richer characters (more unique
characters and mentions) and more accurate character chains (higher
coreference scores).
Although GPT-4V generates more characters than
\textsc{SOtter}++, it does so at the expense of succinctness and
readability, whereas \textsc{SOtter}++ produces stories with lengths
and readability scores close to those of human-written
stories.

\paragraph{LLM-based models improve story diversity.}
We observe that LLM-based models (i.e.,~\textsc{Otter} variants,
\textasciitilde 9B parameters; GPT-4V, undisclosed parameter count)
generate more diverse stories compared to MCSM and Iter-Blueprint,
which are based on smaller language models like BART (\textasciitilde
300M parameters).  This indicates that the extensive prior knowledge
of LLMs prevent them from overfitting on the VIST dataset, which would
otherwise result in similar stories for different images.
Additionally, \textsc{SOtter}++ creates more diverse stories than
\textsc{SOtter}. Character coreference annotations help the model
generate stories that are more target at images provided, avoiding
repetition.
{Table~\ref{tab:aaadditional_results} further demonstrates that \textsc{SOTTER++} achieves better diversity while maintaining the best coherence performance, compared results achieved by prior work.}


\paragraph{Existing grounding metrics fall short of evaluating
  character grounding.}  \textsc{SOtter}++ performs best in
character-centric metrics but is slightly worse than \textsc{SOtter}
in terms of grounding (see column GRD in
Table~\ref{tab:generation_results}). Our analysis suggests that this
discrepancy arises because existing grounding metrics, such as
GROOViST~\cite{surikuchi2023groovist} rely on general-purpose entity
extraction and use CLIPScore to evaluate grounding between text and
images. As a result they do not accurately measure the grounding of
character mentions, particularly when it comes to character names and
pronouns~\cite{liu2023detecting}.

\paragraph{Limitations of MAUVE and n-gram metrics.}
As shown in Table~\ref{tab:generation_results}, LLM-based models (most
dramatically GPT-4V) have lower MAUVE scores than models trained from
scratch on VIST.  This is because MAUVE compares the learned
distribution from a story generation model to the distribution of
human-written stories.  Consequently, the prior knowledge acquired
during LLM pre-training may lead to less accurate alignment with the
distribution of human-authored stories, resulting in lower MAUVE.  We
also observe in Table~\ref{tab:generation_results} that MAUVE and
Diversity exhibit an inverse relationship, suggesting that human
stories in VIST may themselves lack diversity, and we would expect
models  more aligned with these stories to have lower diversity.

We observe an analogous pattern for n-gram based metrics (Table~\ref{tab:generation_results}, right block), including \mbox{Blue-4}, RLSum,
METEOR, and CIDER. These metrics underestimate the performance of
LLM-based models compared to models trained from scratch (MCSM,
Iter-Blueprint). This is because they measure how well a model mimics
the human stories in terms of word-overlap, rather than in terms of 
diversity, grounding, or character-centricity.

\paragraph{{GPT-4V tends to generate rare words and complicated stories.}}
{The reading difficulty scores indicate that the stories generated by GPT-4V are more complex, suggesting that GPT-4V tends to use longer sentences and rare words. Even with five demonstrations for in-context learning, GPT-4V still generates more complex and longer stories.
When explicitly instructed in the prompt to generate stories with fewer than 50 words, GPT-4V produces stories that are close to this length on average, but reading difficulty remains substantially higher than the gold standard and other models.
In contrast, fine-tuned models better match the readability of stories written by humans.}


\begin{table}[t]
\centering
\resizebox{.47\textwidth}{!}{%
\begin{tabular}{ccrrr}
\toprule
VCoref & TCoref & \#CHAR & \#MENTS & COREF \\
\midrule
{\usym{2717}}& {\usym{2717}}& 1.91 & 3.15 & 26.61 \\
\textcolor{ForestGreen}{\usym{2713}} & {\usym{2717}}& 1.89 & 3.19 & 27.54 \\
{\usym{2717}}& \textcolor{ForestGreen}{\usym{2713}} & 2.44 & 3.56 & 27.14 \\
\textcolor{ForestGreen}{\usym{2713}} & \textcolor{ForestGreen}{\usym{2713}} & \textbf{2.98} & \textbf{4.98} & \textbf{32.16} \\
\bottomrule
\end{tabular}%
}
\caption{Ablation study assessing the contribution of different
  annotation components. VCoref stands for visual character
  coreference and TCoref, for textual  character coreference. }
\label{tab:ablation}
\end{table}

\paragraph{Visual coreference alone does not improve character quality.}
Table~\ref{tab:ablation} assesses through an ablation study the impact
of visual character ing and textual coreference.  When we only
annotate the identities of characters in the input images, we do not
observe substantial improvements in the numbers of characters, their
mentions, or the accuracy of coreference in the generated stories
(compare the first and second row in Table~\ref{tab:ablation}).  This
show that despite visually indicating which characters are the same in
the images, the output stories remain practically unchanged and the
model is unable to utilize our visual character annotations
effectively.

\paragraph{Textual coreference improves richness of characters.}
When the model is trained on stories annotated with textual
coreference chains, the predicted stories feature richer character
representations and more mentions (see third row in
Table~\ref{tab:ablation}).  This indicates that applying character
annotations to the target data is a more direct and effective
approach, allowing the model to focus more on the use of characters
during generation.  However, we also observe that in this setting, the
model's coreference accuracy does not improve significantly,
indicating that the model is not better at identifying which
characters in the images are the same person.

\paragraph{Visual and textual coreference combined improve
  coreference.}
When combining visual and textual coreference annotations (last row in
Table~\ref{tab:ablation}), we observe a notable increase in
coreference scores, along with a moderate increase in the number of
characters and mentions.  This indicates that character annotations in
both modalities complement each other, enabling the model to recognize
which characters are the same and leading to more accurate
coreference.

\begin{table}[t]
\centering
\resizebox{.5\textwidth}{!}{%
\begin{tabular}{@{}c@{~}|@{}c@{}c@{}c@{}|c@{}c@{}c|@{}c@{}c@{}c@{}}
\toprule
\multicolumn{1}{@{}c@{}}{} & \textsc{SOtter}\small{++} & vs. &  \multicolumn{1}{c}{Iter-BP} &
\textsc{SOtter}\small{++} & vs. &  \multicolumn{1}{c}{\textsc{SOtter}} & \textsc{SOtter}\small{++} & vs. & Human \\ 
Criteria & Win & Lose & Tie & Win & Lose & Tie  & Win & Lose & Tie \\ \hline
SPE & \textbf{82.3} & 9.5 & 8.2 & \textbf{50.9} & 44.0 & ~~5.2 & 41.0 & \textbf{53.2} & 5.8\\
COH & \textbf{81.5} & 13.1 & 5.5 & 41.6 & \textbf{47.8} & 10.6 & 42.1 & \textbf{51.0} & 6.9\\
ENG & \textbf{64.9} & 26.1 & 9.1 & \textbf{42.3} & 42.2 & 15.4 & 36.0 & \textbf{56.2} & 7.7\\
GRD & \textbf{71.8} & 20.6 & 7.6 & \textbf{57.4} & 40.3 & ~~2.3 & 37.9 & \textbf{56.0} & 6.0\\
CHA & \textbf{88.4} & 8.0 & 3.7 & \textbf{75.1} & 18.2 & ~~6.7 & \textbf{51.6} & 46.8 & 1.6\\ \hline
OVR & \textbf{75.0} & 22.0 & 3.0 & \textbf{70.1} & 22.1 & ~~7.8 & 36.9 & \textbf{55.0} & 8.1\\ \bottomrule
\end{tabular}%
}
\caption{Pair-wise comparison evaluation using \textsc{GPT-4o} as a
  judge. We report the percentage of times \textsc{Otter}++ Wins,
  Loses, or Ties with a comparison system. We evaluate story
  specificity (SPE), coherence (COH), engagement (ENG), grounding
  (GRD), characters (CHA), and overall (OVR).}
\label{tab:llm-pair-wise-results}
\end{table}


\paragraph{\textsc{SOtter}++ leads in side-by-side evaluation.}


Table~\ref{tab:llm-pair-wise-results} summarizes pair-wise comparison
results using \textsc{GPT-4o} as an evaluator. Specifically, we
compare ~\textsc{SOtter}++ against (a)~the iterative Blueprint model
(Iter-BP); (b)~\textsc{SOtter} trained on VIST without character annotations; and (c)~gold-standard stories written by humans.  {We observe
that \textsc{SOtter}++ outperforms the iterative Blueprint model
across all metrics and is superior to \textsc{SOtter} across most
dimensions (with the exception of coherence).}  When compared to
gold-standard stories written, \textsc{SOtter}++ achieves a win rate
of 36.9\%, and is better at generating stories with explicit and
recurring characters (see characters metric).  This suggests that even
human-written stories in the test set tend to use vague plural
pronouns and avoid clear and recurring character usage.

\paragraph{{Generalization performance on the VWP dataset.}}
{Table~\ref{tab:generation_results-VWP} presents the results of evaluating our models on the VWP dataset, when trained on VIST. Since VWP contains more images per story than VIST, the average story length is correspondingly longer. Our results show that \textsc{SOtter}++ demonstrates the strongest generalization ability, outperforming all other models in character-related metrics, grounding, and diversity.}

\begin{figure*}[t]
    \centering
    \includegraphics[width=\textwidth]{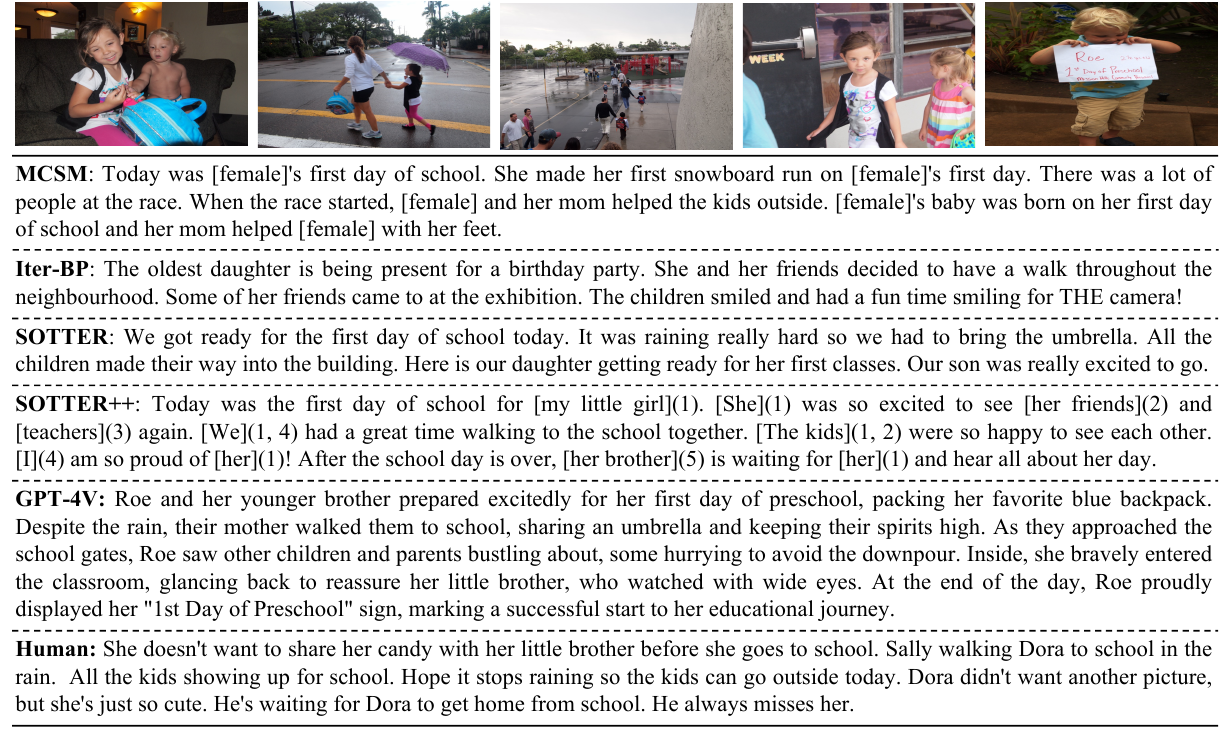}
    \vspace*{-4ex}
    \caption{Examples of system output and human-written story for an
      image sequence (VIST test set).}
    \label{fig:case}
\end{figure*}

\paragraph{\textsc{SOtter}++ generates character-centric stories with
  consistent mentions.}
Figure~\ref{fig:case} shows example stories created by the models used
in our evaluations, as well as by GPT-4V.  We observe that previous
state-of-the-art models, such as MCSM, have poor character mention
capabilities, replacing character names with [female] or [male].
While the stories generated by Iter-BP exhibit more human-like
language, they contain hallucinations, as evidenced by the low
Grounding scores in Table~\ref{tab:generation_results}.  Additionally,
most characters in the story are generic (e.g.,~friends), lacking
explicit mentions to recurring individuals.

In the stories produced by \textsc{SOtter}++, we observe recurring
characters represented by consistent coreference cluster labels.  Even
for plural characters (denoted by ``they'', for instance),
\textsc{SOtter}++ predicts the specific individuals they refer to.
Furthermore, the story language is similar to the style of the human
story.  GPT-4V stories have the highest reading difficulty (RDD; see
Table~\ref{tab:generation_results}): They are long, use rare words,
and employ a flamboyant style which differs from how humans write
visual stories.
In contrast, fine-tuned models better match the readability level of
human-authored stories. There is also a significant error in the story
generated by GPT-4V in Figure~\ref{fig:case}: It incorrectly assumes
that Roe is holding a sign (last story sentence), whereas the image
depicts her younger brother.  {This suggests that even powerful LVLMs like GPT-4V may not reliably utilize character identity across a sequence of images, either due to limitations in their ability to recognize the same character or because they do not consistently attend to this information as a meaningful feature.}


\section{Related Work}
\label{vcoref-related_work}

\paragraph{Visual storytelling}
\citet{huang2016visual} proposed visual storytelling as a way to
create AI systems that can comprehend event sequences and generate
narrative language which goes beyond just describing what is shown in
images.  Early methods
\cite{gonzalez2018contextualize, kim2018glac} used simple
encoder-decoder models with CNNs for visual features and RNNs for text
generation.  Recent approaches
\cite{wang-etal-2024-sco,chen2021commonsense,hsu2020knowledge,yang2019knowledgeable}
leverage external knowledge resources like ConceptNet to enhance
commonsense reasoning abilities.  Some methods
\cite{lu2016visual,hong2020diverse,wang2020storytelling} also utilize
scene graphs to model object relationships.


Few approaches have focused on character-centric story generation.
\citet{surikuchi2019character} extract characters from VIST, analyze
their relationships and exploit them for focusing attention to
relevant visual segments during story generation. However, neither
character coreference nor character grounding are explicitly modeled.
\citet{liu2023detecting} introduce the VIST-Character dataset with
rich character and coreference annotations, however due to its small
size, it can only serve as a test set.

Most existing approaches
\cite{xu2021imagine,hsu2020knowledge,wang2020storytelling,yang2019knowledgeable}
train Transformer models from scratch, with the exception of
\citet{liu2023visual} and \citet{chen2021commonsense}, who employ a
vanilla BART model as a baseline without task-specific adaptation.  In
contrast, our work leverages the language modeling and generalization
capabilities of LLMs (i.e.,~larger than 1B parameters) aiming to
generate visually grounded and character-centric stories.

\paragraph{Character coreference in visual stories}
Previous visual coreference methods
\cite{schroff2015facenet,liu2023detecting} rely on clustering
algorithms which mostly depend on facial features and require a
predefined number of clusters. Facial features often fail to detect
side-view faces, which are common in visual stories, and since
deciding the number of clusters is non-trivial, existing methods often
identify different characters in an image as the same person.  Our
approach eschews these issues by focusing on segmentation masks and
adopting an incremental algorithm which does not require the number
clusters to be known in advance. \citet{hong-etal-2023-visual-writing}
create Visual Writing Prompts (VWP), a dataset which contains visual
stories based on movie shot sequences and manual annotations of
characters aligned to images but no coreference chains. They also
demonstrate that character prompts lead to more diverse and visually
grounded stories. However, in their approach grounding is implicit,
whereas we force the model to align textual mentions to visual
detections via markdown tags.

Current state-of-the-art textual coreference resolution models
generally struggle to effectively handle the coreference of plural
nouns~\cite{liu2023detecting,le2023large,hicke2024lions}, which are
very common in visual storytelling.  Our LLM-based prompting approach
can effectively handle plural and collective nouns and resolve them to
unique characters, overcoming the limitations of existing methods.


\section{Conclusions}

In this work, we proposed the new task of character-centric story
generation and introduced a new dataset, VIST++, which extends the
widely use VIST dataset with visual and textual character grounding
and coreference. The dataset comprises annotations for 300K unique
characters over 40K visual stories.  We then described a new model for
character-centric story generation; this model was finetuned on VIST++
and built on \textsc{Otter}, a pre-trained large vision and language
model. Our evaluation showed the richness of the characters and the
accuracy of grounding and coreference in the stories our model
generates. Furthermore, we propose an LLM-as-a-judge evaluation which
demonstrates that stories generated by our model are preferred over
stories generated without character grounding and coreference.

\paragraph*{Limitations}
It is worth noting that more visual storytelling datasets have emerged
recently \cite{ravi2021aesop,hong-etal-2023-visual-writing}.  However,
given that VIST remains the most widely used dataset for visual
storytelling and considering the costs of annotation and
experimentation, our work focused on VIST. Although the proposed
character-centric method improves the generated stories, there is
still room for improvement in terms of the coreference scores
achieved; future work focused on improving coreference is likely to
further improve story generation.

\bibliography{custom}
\bibliographystyle{acl_natbib}

\onecolumn

\appendix

\section{Character Clustering Algorithm}
\label{app:sequential_matching}


In this section we formally present our incremental clustering
algorithm. We also explain how we compute pairwise
character matching with a QA method based on LVLMs  which we experimentally
found to be superior to the more widely used cosine similarity of visual
features.

Specifically, we provide crops of two characters to a LVLM and ask
whether they refer to the same individual.  Since character crops
often include background noise (e.g., other characters within the
bounding box, such as two characters hugging), we employ fine-grained
visual prompting \cite{yang2024fine} to focus the model's attention on
the primary character in the crop.  We compute visual similarity using
\textsc{LlaVA1.5-13b} \cite{liu2024llavanext}. As this model does not
support multi-image input, we concatenate the images of the two
characters into a single image. The prompt we used is: \textit{Are the person on the left and the person on the right the same? Output yes or no ONLY.}



\begin{algorithm}[t]
\caption{Incremental Character Matching}
\label{alg:sequential_character_matching}
\textbf{Input}: Image sequence $\{I_i\}_{i=1}^{K}$\\
\textbf{Output}: Visual coreference chains $\{\mathcal{V}_i\}_{i=1}^{N_v}$ \textbf{Start:}
\begin{algorithmic}[1] 
\STATE Initialize visual coref chains $\{\mathcal{V}_i\}_{i=1}^{N_v} \leftarrow \emptyset$
\STATE Detect characters in $I_1$ using DETR, obtain detections $D_1$
\FOR{each detection $d \in D_1$}
    \STATE Add $d$ to a new chain in $\{\mathcal{V}_i\}$
\ENDFOR

\FOR{$i = 2$ to $K$}
    \STATE Detect characters in $I_i$ using DETR, obtain detections $D_i$
    \STATE Extract visual features of $D_i$ using CLIP
    \STATE Initialize similarity matrix $M_{q \times p}$, where $q = |D_i|$ and $p = |\{\mathcal{V}_i\}|$
    \FOR{each detection $d_q \in D_i$}
        \FOR{each existing character $c_p \in \{\mathcal{V}_i\}$}
            \STATE Compute matching score $M[q, p]$ between $d_q$ and $c_p$ using CLIP features
        \ENDFOR
    \ENDFOR
    \STATE Solve Bipartite Graph Matching problem using Hungarian algorithm on $M$
    \FOR{each detection $d_q \in D_i$}
        \IF{$d_q$ matches to existing character $c_p$}
            \STATE Add $d_q$ to the chain of $c_p$
        \ELSE
            \STATE Create a new chain in $\{\mathcal{V}_i\}$ for $d_q$
        \ENDIF
    \ENDFOR
\ENDFOR

\STATE \textbf{return} Visual coreference chains $\{\mathcal{V}_i\}_{i=1}^{N_v}$
\end{algorithmic}
\end{algorithm}

\section{LLM-as-a-Judge Details}
\label{app-llmjudge}

We utilized a collection of human evaluation results from four VIST studies: KG-Story~\cite{hsu2020knowledge}, PR-VIST~\cite{hsu2021plot}, Streth-VIST~\cite{hsu2021stretch}, and Iter-Blueprint~\cite{liu2023visual}.
Specifically, the first four datasets include annotations for side-by-side overall-level comparisons of two stories, from which we randomly selected 1,000 story pairs. In the Iter-Blueprint dataset, we used the full set of fine-grained annotations for 100 stories, covering aspects such as coherence, engagement, grounding, and overall quality.

Additionally, the authors randomly selected 100 story pairs from Iter-Blueprint and VIST gold stories for manual annotation, focusing on specificity and character-based side-by-side comparisons. The criteria of each of these metrics can be found in Table~\ref{tab:prompt_quality_estimation2}.

\section{List of Prompts}
{This section provides all prompts used in this study. Table~\ref{tab:prompt_character_detection} presents the zero-shot prompt used for detecting character words. Table~\ref{tab:prompt_gpt4v} provides the zero-shot prompt used for generating visual stories using GPT-4. Table~\ref{tab:prompt_coreference_resolution} contains the five-shot prompt used for character coreference resolution. Finally, Table~\ref{tab:prompt_quality_estimation2} displays the zero-shot prompt used for pairwise story comparison.}

\begin{table*}[t!]\centering
\begin{minipage}{1.0\columnwidth}\vspace{0mm}    \centering
\begin{tcolorbox} 
\centering
\footnotesize
\begin{tabular}{lp{0.9\columnwidth}}
\multicolumn{2}{l}{Read the following short story and answer the question to identify
words referring to people, which includes:} \\
\\
1 .&   Named entities, including both personal names (e.g.,~`John', `Alice') and specific designations (e.g.,~`President', `CEO').\\
2. & Pronouns as subjects (e.g.,~he, she, they, I, we) and objects (e.g.,~him, her, them, us), but excluding possessive forms (e.g.,~his, our, my, mine).\\
3. & Terms denoting groups of people, including team names, familial
terms, and other plural or compound nouns (e.g.,~`the Smith family',
`committee', `USA team'). 
\\ \\
\multicolumn{2}{l}{\VarSty{ \bf Context: } \{\{\textit{context}\}\}}\\
\\
\multicolumn{2}{l}{\VarSty{ \bf Question:}}
\\
\multicolumn{2}{l}{In the sentence \{\{\textit{sent}\}\}, does the word \{\{\textit{noun}\}\} refer to people?}\\
\\
\multicolumn{2}{l}{\VarSty{ \bf Answer (yes or no ONLY):}}\\

    \end{tabular}
\end{tcolorbox}
\vspace{-2mm}
\caption{Zero-shot prompt used for detecting character
  words. ``\{\{\textit{context}\}\}'', ``\{\{\textit{sent}\}\}'', and
  ``\{\{\textit{noun}\}\}'' are replaced by the story, the sentence
  that the query words are in, and the query words, respectively.}
\label{tab:prompt_character_detection}
\end{minipage}
\end{table*}

\begin{table*}[t!]\centering
\begin{minipage}{1.0\columnwidth}\vspace{0mm}    \centering
\begin{tcolorbox} 
\centering
\footnotesize
\begin{tabular}{lp{0.9\columnwidth}}
\multicolumn{2}{l}{\VarSty{ \bf Image Sequence: } \{\{\textit{Image Sequence}\}\}}\\
\\
\multicolumn{2}{l}{Read the image sequence and write a 5-sentence short coherent story based on the image sequence.} \\
    \end{tabular}
\end{tcolorbox}
\vspace{-2mm}
\caption{Zero-shot prompt used for generating visual stories using GPT-4V. ``\{\{\textit{Image Sequence}\}\}'' will be replaced by the input image sequence.}
\label{tab:prompt_gpt4v}
\end{minipage}
\end{table*}

\begin{table*}[t!]\centering
\begin{minipage}{1.0\columnwidth}\vspace{0mm}    \centering
\begin{tcolorbox} 
\centering
\footnotesize
\begin{tabular}{p{0.97\columnwidth}}
Annotate all the people mentions within square brackets in the following text with coreference clusters. Use Markdown tags to indicate clusters in the output, with the following format [mention](\#cluster\_name). For plural nouns, include all related clusters. \\
\\
Input:\\
 
[My friends] and [I] went on a history trip. On the boat, [we] took pictures of other boats like this. [We] used this rope to bring in the anchor for our boat. [One of my partners] posed on the ship for a picture. While the sun went down, [we] cruised and took pictures of the city line.\\

Output:\\
 
[My friends](\#1) and [I](\#2) went on a history trip. On the boat, [we](\#1, \#2) took pictures of other boats like this. [We](\#1, \#2) used this rope to bring in the anchor for our boat. [One of my partners](\#1) posed on the ship for a picture. While the sun went down, [we](\#1, \#2) cruised and took pictures of the city line.\\
\\
Input:\\
 
It is date night for [Tom] and [Susan]. Being sports fans [they] decided to attend a baseball game. [They] are having a great time and being goofy. [Tom] strikes a pose with his sweet ride. Meanwhile, [their kids] are at home with the [babysitter].\\

Output:\\
 
It is date night for [Tom](\#1) and [Susan](\#2). Being sports fans [they](\#1, \#2) decided to attend a baseball game. [They](\#1, \#2) are having a great time and being goofy. [Tom](\#1) strikes a pose with his sweet ride. Meanwhile, [their kids](\#3) are at home with the [babysitter](\#4).\\
\\
Input:\\
 
[James] and [I] were excited to be in Washington D.C. during the 4th of July. There was [a huge crowd of people] already awaiting the firework show. [We] were lucky to find a nice spot on the grass to watch the show. As the evening grew darker [the crowd] was gearing up to enjoy the show, with a great view of the Washington Monument. [I] was able to capture a great photo of the grand finale of the firework show.\\

Output:\\
 
[James](\#1) and [I](\#2) were excited to be in Washington D.C. during the 4th of July. There was [a huge crowd of people](\#3) already awaiting the firework show. [We](\#1, \#2) were lucky to find a nice spot on the grass to watch the show. As the evening grew darker [the crowd](\#4) was gearing up to enjoy the show, with a great view of the Washington Monument. [I](\#2) was able to capture a great photo of the grand finale of the firework show.\\
\\
Input:\\
 
[Grandpa] was happy to have [the family] over for the 4th. [His granddaughter] also came for a visit.  [They] were so happy to see each other. [They] played for a long time. [Grandpa] took [her] to see some fireworks. [She] was scared of the loud boom, but excited by the bright colors. It was a perfect day.\\

Output:\\
 
[Grandpa](\#1) was happy to have [the family](\#2) over for the 4th. [His granddaughter](\#3) also came for a visit.  [They](\#1, \#2, \#3) were so happy to see each other. [They](\#1, \#2, \#3) played for a long time. [Grandpa](\#1) took [her](\#3) to see some fireworks. [She](\#3) was scared of the loud boom, but excited by the bright colors. It was a perfect day.\\
\\
Input:\\
 
Today would be [the baby]'s first trip to the beach. [Daddy] took [the baby] to the beach and showed [him] how the sand and ocean looked. Here the duo are taking a picture snapped by [the mother]. This time [they] all got in on the picture and had a blast! [The baby] had a lot of fun on his first beach trip!\\

Output:\\
 
Today would be [the baby](\#1)'s first trip to the beach. [Daddy](\#2) took [the baby](\#1) to the beach and showed [him](\#1) how the sand and ocean looked. Here the duo are taking a picture snapped by [the mother](\#3). This time [they](\#1, \#2, \#3) all got in on the picture and had a blast! [The baby](\#1) had a lot of fun on his first beach trip!\\
\\
Input:\\
 
\{\{story\}\}\\
\\
Output:
\end{tabular}
\end{tcolorbox}
\vspace{-2mm}
\caption{The 5-shot prompt used for character coreference
  resolution. ``\{\{\textit{noun}\}\}" is replaced by characters
  previously detected in the input story.}
\label{tab:prompt_coreference_resolution}
\end{minipage}
\end{table*}

\begin{table*}[t!]\centering
\begin{minipage}{1.0\columnwidth}\vspace{0mm}    \centering
\begin{tcolorbox} 
\centering
\footnotesize
\begin{tabular}{p{0.97\columnwidth}}
You will be asked to compare two stories generated based on the input images, according to specific criteria. The aim is to assess the quality of each story and select the better one in terms of specificity, coherence, engagement, grounding, characters, and overall preference.
Please follow the guidelines below for your evaluation:    \\
\\
\VarSty{ {\bf Criteria for Evaluation:}} \\

1. \textbf{Specificity}: Check if the story avoids general statements like "we had a good time" or "I enjoyed myself". A good story should provide detailed, specific examples and descriptions rather than relying on broad, vague statements.\\
2. \textbf{Coherence}: Assess if the story flows smoothly, maintaining logical connections and consistency that make senses to the reader.\\
3. \textbf{Engagement}: Evaluate how interesting the story is, taking into account its unique and memorable features.\\
4. \textbf{Grounding}: Determine how closely the story aligns with the provided input images, capturing their key elements.\\
5. \textbf{Characters}: Check whether the story has clear, recurring characters.\\
6. \textbf{Overall Preference}: Choose the preferred story out of the two provided, based on a combination of specificity, coherence, engagement, grounding, characters and an overall preference.\\
\\
\VarSty{ {\bf Evaluation Steps:}} \\
1. Carefully read the two stories to grasp the main theme and key points.\\
2. Compare the two story based on the dimensions provided above (specificity, coherence, engagement, grounding, characters and an overall preference).\\
3. Construct a dictionary in Python format with your preferred story for each criterion, using 'A' if Story A is better or 'B' if Story B is better. Your feedback should look like this:\\
\textit{\{}\\
\;\;\;\;\textit{"specificity": "A" or "B",} \\
\;\;\;\;\textit{"coherence": "A" or "B",} \\
\;\;\;\;\textit{"engagement": "A" or "B",} \\
\;\;\;\;\textit{"grounding": "A" or "B",} \\
\;\;\;\;\textit{"characters": "A" or "B",} \\
\;\;\;\;\textit{"overall\_preference": "A" or "B"} \\
\textit{\}}\\
 \\
\VarSty{ {\bf Stories:}}  \\
 \\
\textit{Story A: \{\{story\_a\}\}} \\
 \\
 \textit{Story B: \{\{story\_b\}\}} \\
 \\
\VarSty{ {\bf Output (dictionary ONLY):}}  \\

    \end{tabular}
\end{tcolorbox}
\vspace{-2mm}
\caption{The zero-shot prompt used for pair-wise story comparison. "\textit{\{\{story\_a\}\}}" and "\textit{\{\{story\_b\}\}}" will be replaced by stories generated by our systems or other models.}
\label{tab:prompt_quality_estimation2}
\end{minipage}
\end{table*}

\section{VIST++ Dataset: Quality Evaluation}
\label{app:pipeline_anylasis}


In this section we analyze the quality of our automated annotation pipeline. Recall  
that we detect  character bounding boxes  with DETR~\cite{carion2020end} and use OpenCLIP
ViT-G/14~\cite{ilharco_gabriel_2021_5143773} for feature extraction.
\textsc{LLaMA3-70B} is used for textual coreference, and SAM
\cite{kirillov2023segment} is employed to obtain the segmentation
masks for bounding boxes.  For LVLM-based pair-wise character matching,
we employ \textsc{LLaVA1.5-13b}~\cite{liu2024llavanext}. We evaluate our pipeline against  
manual annotations provided in VIST-Character \cite{liu2023detecting} as well as alternative
implementations of character detection and coreference modules. 


Our results are summarized in Table~\ref{tab:detec-coref-results}.
For textual and visual character detection, we report precision and recall.
We assume a textual detection is correct if the head of the noun phrase
in question is the same as the gold-standard one.
For visual detections, a predicted region is considered correct when the Intersection over 
Union (IoU; \citealt{yu2016unitbox} is higher than a threshold of 50\%. 
For character coreference, we evaluate the coreference chains using B$^3$ metrics~\citet{cai2010evaluation}, which indicate the average percentage of the correctly detected mentions in a chain.
For each mention, the B$^3$ algorithm computes a precision and recall score using the following equations:
\begin{equation}
\begin{aligned}
\operatorname{Precision}\left(m_i\right) & =\frac{\left|R_{m_i} \cap G_{m_i}\right|}{\left|R_{m_i}\right|} \\
\operatorname{Recall}\left(m_i\right) & =\frac{\left|R_{m_i} \cap G_{m_i}\right|}{\left|G_{m_i}\right|}
\end{aligned}
\end{equation}
where $R_{m_i}$ is the chain from the system prediction, which includes the mention $m_i$ , and $G_{m_i}$ is the manually annotated gold-standard chain with $m_i$.
The overall precision and recall are computed by averaging them over all mentions.
For multimodal alignment, we use recall and precision.

\begin{table}[t]
\centering
\resizebox{.47\textwidth}{!}{%
\begin{tabular}{lcccc}
\hline
\multicolumn{1}{c}{} & \multicolumn{2}{c}{Detection} & \multicolumn{2}{c}{Coreference} \\
\multicolumn{1}{l}{\multirow{-2}{*}{Model}} & P & R & P & R \\
\hline
WordNet + SpanBERT & 74.0 & 90.3 & 66.6 & 70.0 \\
\textbf{VIST++} & \textbf{84.9} & \textbf{94.5} & \textbf{77.2} & \textbf{80.3} \\
\hline
CLIP + Kmeans & 40.5 & 69.1 & 51.8 & 60.8 \\
\textbf{VIST++} & \textbf{41.8} & \textbf{78.4} & \textbf{70.4} & \textbf{74.2} \\ \hline
\end{tabular}
}
\caption{Results for textual (upper part) and visual (lower part) character detection and coreference. P and R are short for precision and recall.
}
\label{tab:detec-coref-results:appendix}
\end{table}




The first block  in Table~\ref{tab:detec-coref-results} presents results for textual character detection and coreference resolution, while the second block does the same for visual characters. 
We observe that our QA-prompting method brings a significant improvement in the precision of textual character mention detection.
Previous methods \cite{liu2023detecting,surikuchi2019character} employing WordNet to filter character words often lead to false positives.
This is because they   operate at the word/phrase level and do not consider contextual semantic information. For example, ``white'' is usually used to describe color, but its primary sense in WordNet is a person’s name. Also, some character words like ``great-grandmother'' are not included in WordNet.
However, the strong contextual understanding and reasoning capabilities of LLMs can resolve these issues effectively in a question-answering setting.

\begin{table}[t]
\centering
\resizebox{.40\textwidth}{!}{%
\begin{tabular}{lcc}
\hline
Model & Recall & Precision \\ \hline
\citet{liu2023detecting} & 27.3 & 27.5 \\
\textbf{Our Pipeline} & \textbf{39.7} & \textbf{42.5}\\ \hline
\end{tabular}%
}
\caption{Performance of  mutlimodal coreference chain alignment. All numbers are  percentages (\%).}
\label{tab:alignment_result}
\end{table}

Our prompting-based method shows approximately a 10\% improvement in precision and recall for character coreference.
This demonstrates that LLMs are effective coreference resolvers and that structured prompting can predict all sub-clusters referred to by plural nouns, a task that previous methods based on SpanBERT struggled with.
It is worth noting that in our experiments, only large LLMs  were capable of handling coreference tasks (we used \textsc{LLaMA3-chat-70B}).
Smaller models (e.g., \textsc{LLaMA3-chat-8B}) often failed to understand the complex structure of the prompts, resulting in outputs that did not follow instructions. 


Our incremental clustering algorithm is superior to  previous  clustering approaches based on facial features. 
Using body features reduces the number of missing characters, especially when their faces are not visible in the images.
Additionally, our algorithm addresses a critical limitation of previous clustering approaches: the same person cannot appear more than once in a single image.
These improvements effectively enhance the accuracy of visual character coreference, resulting in a 10\%--15\% increase in the accuracy of multimodal chain alignment, as shown in Table~\ref{tab:alignment_result}.
Overall, our  results demonstrate that our pipeline achieves state-of-the-art performance in character coreference resolution.




\label{sec:appendix} 

\end{document}